\documentclass[conference]{IEEEtran}
\IEEEoverridecommandlockouts
\usepackage{amsmath,amssymb,amsfonts}
\usepackage{algorithmic}
\usepackage{graphicx}
\usepackage{textcomp}
\usepackage{xcolor}
\def\BibTeX{{\rm B\kern-.05em{\sc i\kern-.025em b}\kern-.08em
    T\kern-.1667em\lower.7ex\hbox{E}\kern-.125emX}}
    
\makeatletter
\newcommand{\linebreakand}{%
  \end{@IEEEauthorhalign}
  \hfill\mbox{}\par
  \mbox{}\hfill\begin{@IEEEauthorhalign}
}
\makeatother

\usepackage{bbm}

\usepackage{amsmath,amsfonts,bm}
\usepackage{amssymb}
\usepackage{amsthm}
\usepackage{mathtools} 

\def\eqref#1{equation~\ref{#1}}

\def\1{\bm{1}}

\def\mD{{\bm{D}}}

\def\mI{{\bm{I}}}

\def\mK{{\bm{K}}}
\def\mL{{\bm{L}}}

\def\mP{{\bm{P}}}

\DeclareMathAlphabet{\mathsfit}{\encodingdefault}{\sfdefault}{m}{sl}
\SetMathAlphabet{\mathsfit}{bold}{\encodingdefault}{\sfdefault}{bx}{n}

\usepackage{booktabs} 

\begin{document}
\title{MURAL: An Unsupervised Random Forest-Based Embedding for Electronic Health Record Data\\
\thanks{
* Equal contribution. $\dagger$ Corresponding author. This research was partially supported by IVADO Professor funds, CIFAR AI Chair, and NSERC Discovery grant 03267 [\emph{G.W.}]; Chan-Zuckerberg Initiative grants 182702 \& CZF2019-002440 [\emph{S.K.}]; and NIH grants R01GM135929 \& R01GM130847 [\emph{G.W., S.K.}], K23DK125718-01A1 [{\emph{D.S.}]}. The content provided here is solely the responsibility of the authors and does not necessarily represent the official views of the funding agencies.}
}

\author{\IEEEauthorblockN{Michal Gerasimiuk$^*$}
\IEEEauthorblockA{\textit{Dept. of Comp. Sci.} \\
\textit{Yale University} \\
New Haven, CT, USA \\
michal.gerasimiuk@yale.edu}

\and 

\IEEEauthorblockN{Dennis Shung$^*$}
\IEEEauthorblockA{\textit{Sch. of Medicine} \\
\textit{Yale University} \\
New Haven, CT, USA \\
dennis.shung@yale.edu}

\and

\IEEEauthorblockN{Alexander Tong}
\IEEEauthorblockA{\textit{Dept. of CS \& Oper. Res.} \\
\textit{Univ. de Montr\'{e}al; Mila} \\
Montreal, QC, Canada \\
alexander.tong@mila.quebec}

\linebreakand

\IEEEauthorblockN{Adrian Stanley}
\IEEEauthorblockA{Glasgow Royal Infirmary\\
Glasgow G4 0SF, United Kingdom\\
adrianstanley99@gmail.com
}

\and 

\IEEEauthorblockN{Michael Schultz}
\IEEEauthorblockA{\textit{Department of Medicine}\\
\textit{Otago Med. Sch. -– Dunedin Campus}\\
Dunedin, New Zealand\\
michael.schultz@otago.ac.nz
}

\and 

\IEEEauthorblockN{Jeffrey Ngu}
\IEEEauthorblockA{\textit{Christchurch Hospital}\\
Christchurch, New Zealand\\
jeffrey.ngu@cdhb.health.nz}

\linebreakand

\IEEEauthorblockN{Loren Laine}
\IEEEauthorblockA{\textit{Sch. of Medicine} \\
\textit{Yale University} \\
New Haven, CT, USA \\
loren.laine@yale.edu}

\and 

\IEEEauthorblockN{Guy Wolf}
\IEEEauthorblockA{\textit{Dept. of Math. and Stat.} \\
\textit{Univ. de Montr\'{e}al; Mila} \\
Montreal, QC, Canada \\
guy.wolf@umontreal.ca}

\and 

\IEEEauthorblockN{Smita Krishnaswamy$^\dagger$}
\IEEEauthorblockA{\textit{Depts. of Genetics \& Comp. Sci.} \\
\textit{Yale University} \\
New Haven, CT, USA \\
smita.krishnaswamy@yale.edu
}}

\IEEEoverridecommandlockouts
\IEEEpubid{\makebox[\columnwidth]{978-1-6654-3902-2/21/\$31.00~\copyright2021 IEEE \hfill} \hspace{\columnsep}\makebox[\columnwidth]{ }}

\maketitle

\begin{abstract}
A major challenge in embedding or visualizing clinical patient data is the heterogeneity of variable types including continuous lab values, categorical diagnostic codes, as well as missing or incomplete data. In particular, in EHR data, some variables are {\em missing not at random (MNAR)} but deliberately not collected and thus are a source of information. For example, lab tests may be deemed necessary for some patients on the basis of suspected diagnosis, but not for others. Here we present the MURAL forest -- an unsupervised random forest for representing data with disparate variable types (e.g., categorical, continuous, MNAR). MURAL forests consist of a set of decision trees where node-splitting variables are chosen at random, such that the marginal entropy of all other variables is minimized by the split. This allows us to also split on MNAR variables and discrete variables in  a way that is consistent with the continuous variables. The end goal is to learn the MURAL embedding of patients using average tree distances between those patients. These distances can be fed to nonlinear dimensionality reduction method like PHATE to derive visualizable embeddings. While such methods are ubiquitous in continuous-valued datasets (like single cell RNA-sequencing) they have not been used extensively in mixed variable data. We showcase the use of our method on one artificial and two clinical datasets. We show that using our approach, we can visualize and classify data more accurately than competing approaches. Finally, we show that MURAL can also be used to compare cohorts of patients via the recently proposed tree-sliced Wasserstein distances. 

\end{abstract}


\section{Introduction}

Unsupervised nonlinear embedding methods have allowed for exploration manifold learning of big high dimensional datasets in many fields ranging from epidemiology, to biology, to physics. However, a major limitation of using unsupervised embeddings in healthcare data is the large amount of missingness in the data as well as the mixed modality of the variables collected. In a typical EHR or patient dataset the range of missing data range from 20\% to 80\%, varying across broad categories of possible fields such as demographics, laboratory values, and treatment information~\cite{kopcke2013,hu2017,groenwold2020}. Further there is a mix of real-valued, categorical and binary data which can be difficult to normalize or scale. This makes it difficult to to compute distances and affinities between datapoints---the first step in nonlinear dimensionality reduction methods such as  tSNE~\cite{vanderMaaten2008tSNE}, UMAP~\cite{mcinnes2018umap}, diffusion maps~\cite{coifman:diffusionMaps2006} or PHATE~\cite{Moon19}. Similar distance/affinity computations are also required for spectral clustering~\cite{ng2001spectral}, which operates on a graph Laplacian computed from the affinity matrix. Thus data with missing values cannot be used, and if the values are MNAR they cannot be imputed.

To tackle these issues, we propose to use an intermediary representation called the MURAL-forest, an unsupervised random forest in which tree distances between datapoints form an accurate measure of dissimilarity and can be used for data distance/affinity computation, as needed in \cite{vanderMaaten2008tSNE, Moon19, mcinnes2018umap, coifman:diffusionMaps2006}.  MURAL creates a set of trees by splitting on any variable type (categorical, continuous, with or without missingness) using a marginal entropy criterion that is computed on {\em other} variables. Further, MURAL ensures that heterogeneity within categorical or MNAR variables is immediately broken down using low dimensional entropy to create 4-way splits at such levels. We test MURAL on ground truth data that the resulting tree distances result in accurate embeddings. 

While random forests are normally supervised and trained for prediction, there have been some efforts to learn random forests in an unsupervised manner. \cite{criminisi2012decision} describes a method called {\em manifold forests} which effectively use a splitting criterion based on intra-versus-inter split affinity or density. However, these and other methods often presuppose the ability to compute distances or affinities between high dimensional datapoints. By contrast, we use our MURAL unsupervised random forests in order {\em to be able to compute} an accurate distance between datapoints with missing and mixed-mode variables. 

We show the accuracy of our method by comparing the MURAL derived distances to known ground truth and recovering embeddings in a 5-dimensional Swiss roll. We then apply our method to a complete case subset of an intensive care unit dataset and of an international patient registry dataset of patients presenting with symptoms of upper gastrointestinal bleeding. We induce missingness in the complete case subsets in specific ranges of laboratory values and compare imputed values using mean imputation and multiple imputation with chained equations to the original ground truth.  We then construct MURAL-embeddings on the full datasets with missingness. We show that MURAL-embeddings consistently display more structure and create separations that are more clinically meaningful than commonly used imputation methods.  Finally, we show an application of our method in comparing entire cohorts of patients by computing a tree-based Wasserstein distance on the MURAL-forest, which can be used to quantify similarities or distances between patient cohorts.

 
\section{Background}
\subsection{Manifold Learning, Dimensionality Reduction, Clustering}

Though there are many nonlinear dimensionality reduction and embedding methods, we focus our results on methods that can learn the {\em data manifold} or intrinsic low dimensional shape and structure of the data. We believe that this is useful in biomedical settings where many measurements of the patient reflect non-orthogonal aspects of the same underlying entity, essentially indicating the data in fact lies in a lower dimensional space. 

High dimensional data can often be modeled as a sampling $Z = \{z_i\}_{i=1}^N \subset \mathcal{M}^d$ of a $d$ dimensional manifold $\mathcal{M}^d$ that is mapped to $n$ dimensional observations $X = \{x_1, \ldots, x_N\} \subset \mathbbm{R}^n$ via a nonlinear function $x_i = f(z_i)$. Intuitively, although measurement strategies, modeled here via $f$, create high dimensional observations, the intrinsic dimensionality, or degrees of freedom within the data, is relatively low. This manifold assumption is at the core of the vast field of manifold learning \cite{moon2018manifold,coifman:diffusionMaps2006, van2009dimensionality, izenman2012introduction}, which leverages the intrinsic geometry of data, as modeled by a manifold, for exploring and understanding patterns, trends, and structure that displays significant nonlinearity.

In \cite{coifman:diffusionMaps2006}, diffusion maps were proposed as a robust way to capture intrinsic manifold geometry in data by eigendecomposing a powered diffusion operator. Using $t$-step random walks that aggregate local affinity, \cite{coifman:diffusionMaps2006} were able to reveal nonlinear relations in data and allow their embedding in low dimensional coordinates. These local affinities are commonly constructed using a Gaussian kernel:
\begin{equation}
\label{GKernel}
\mathbf{K} (x_i, x_j) = \exp\left( {-\frac{\| x_i- x_j\|^2}{\varepsilon}  }\right) \,, \quad i,j=1,...,N
\end{equation}
where $\mK$ forms an $N \times N$ Gram matrix whose $(i,j)$ entry is denoted by $\mK(x_i, x_j)$. A diffusion operator is defined as the row-stochastic matrix $\mP = \mD^{-1} \mK$ where $\mD$ is a diagonal matrix with $\mD (x_i, x_i) = \sum_j \mK (x_i,x_j)$. The matrix $\mP$, or diffusion operator, defines single-step transition probabilities for a time-homogeneous diffusion process, or a Markovian random walk, over the data. Furthermore, as shown in \cite{coifman:diffusionMaps2006}, powers of this matrix $\mP^t$, for $t > 0$, can be used to simulate multi-step random walks over the data, helping understand multiscale organization of $X$, which can be interpreted geometrically when the manifold assumption is satisfied. $\mP$ has been used in many downstream unsupervised learning tasks, eigendecomposition of $\mP$ yields the popular diffusion map dimensionality reduction method that can be used as input to clustering. $\mP$ is also used by the PHATE~\cite{Moon19} method for visualization. PHATE transforms the diffusion operator with a pointwise logarithm $\log(\mP)$, derives distances between points $x_i, x_j$ as $\|\log(\mP_i)-\log(\mP_j)\|_2$, and then embeds the resulting distances, known as {\em potential distances}, with metric MDS.

Other methods for visualization, such as \cite{vanderMaaten2008tSNE, mcinnes2018umap}, use $\mK$ rather than $\mP$ to focus on near neighbors rather than learning the entire data manifold. The diffusion operator $\mP$ is related to the graph Laplacian that, depending on the normalization used, can be written as $\mL=\mI-\mK$ or $\mL=\mI-\mP$. Thus the graph Laplacian has the same eigenvectors and eigenvalues that are in the opposite order. Spectral clustering~\cite{ng2001spectral} is often described in terms of the graph Laplacian, i.e., $k$-means over a graph Laplacian rather than data.  
\subsection{Decision Trees}
\label{sec:decisiontree}

A {\em tree} $T$ is a connected directed acyclic graph $T=(V,E)$ with vertices (or nodes) $V=\{t_1, t_2, \ldots, t_n\}$ and $n-1$ edges $E$ such that every node has at most one incoming edge. A rooted tree has a {\em root} node $t_1$ with no incoming edges, while $t_i$, $i > 1$, all have exactly one incoming edge. A node $t_j \in children(t_i)$ if and only if $[t_i, t_j] \in E$, i.e., there is a directed edge from $t_i$ to $t_j$. A $descendant(t_i)$ is any node $t_k$ that is connected to $t_i$ by a directed path $t_i, \ldots t_k$ width a directed edge between each consecutive pair of nodes.

Decision trees contain nodes that split on a variable to create partitions of the data such that datapoints on one side of the partition are more similar to each other in terms of the decision variable. Recursive splits create finer granularity branches where data points are similar with respect to all of the variables that have been split on the path to the node. A specific strength of decision tree is the ability to naturally split multiple types of data---binary, ordinal, and missing. 

\subsection{Supervised Random Forests}

In classification tasks single decision trees can learn irregular patterns and overfit to data. As a way of addressing this, random forests average over sets of decision trees~\cite{breiman1984classification} and are created by randomizing variable splits. The algorithm selects a random subset of features at each potential split, and chooses a threshold so as to optimize a local criterion such as the {\em Gini impurity index} or {\em information gain}. The Gini impurity index is an information theoretic measure that is based on Tsallis entropy~\cite{tsallis1988entropy}. For $C$ classes (given labels) with fractions $P = \{ p_1, p_2, \ldots, p_C\}$ of observations in each class, the Gini impurity index is given by $I_G(P) = 1- \sum_i p_i$. Information gain is also an information theoretic measure which measures the difference in Shannon entropy between the parent node and child nodes. Shannon entropy of a probability distribution $P$ is given by $H(P) = -\sum_i p_i \log(p_i)$. Information gain is defined as
\begin{equation}
    \label{eqn:infogain}
    I_G(P) = H(P) - \sum_a \frac{|a|}{k} H(P^a).
\end{equation}
Here $P$ is the class distribution of the parent node, and $P^a$ is the class distribution of the $a$-th child node, which receives $|a|$ datapoints. The total number of datapoints split by the parent node is $k$. Note that these criteria are with respect to a classification label that is given in a supervised setting. 

The original random forest classifier used labeled data to randomly train an ensemble of decision trees with a majority vote aggregating the classifications. Decision trees are constructed through recursively partitioning the space occupied by data as observations travel from the tree's root to its leaves, each nonterminal node containing a weak learner that chooses a splitting variable and threshold. These weak learners minimize an impurity function to ensure that each child node receives a ``purer'' cohort than its parent. Purity is determined by the proportion of labels; if all examples belong to the same class, the subset is considered pure. 

\subsection{Unsupervised Random Forests}
Variants of decision trees have been used to cluster data in the absence of labels: random projection trees~\cite{hedge2008projection, dasgupta2008random}, density forests~\cite{criminisi2012decision}, PCA trees~\cite{verma2009spatial}, approximate principal direction trees~\cite{mccartin-lim2012approximate}, and geodesic forests~\cite{madhyastha2020geodesic}. These variants are often effective at learning the manifold of the data when the data variables are continuous and distances or Gaussian affinities can be defined between datapoints. However, for us this creates a chicken-and-egg problem. Our purpose in creating a random forest is to derive a meaningful distance in situations where there are missing values and categorical variables, where simple Euclidean distances are not meaningful.

For example, Criminisi's manifold forests~\cite{criminisi2012decision} use trees whose nodes minimize the following information gain measure when splitting
\begin{equation}
    \label{eqn:Criminisi-infogain}
    I_G(S_j) = \log(| \Lambda(S_j) |) - \sum_{i \in \{L, R\}} \frac{|S_j^i|}{|S_j|} \log(| \Lambda(S_j^i) |).
\end{equation}
Here, $S_j$ is the set of datapoints that node $j$ partitions, $S_j^L$ and $S_j^R$ are the sets of datapoints from $S_j$ that get sent to the left and right child of node $j$, respectively. The matrix $\Lambda(S)$ is a set's covariance matrix, which is undefined in our case with missing values. Furthermore, unless binary affinities are chosen, the affinity matrices defined using manifold forests depend on preexisting distances between datapoints. Thus we define a new type of tree that can tolerate missing values and mixtures of variables, which can itself be used to compute a new type of distance. 

\subsection{Wasserstein Distance over Trees}

The 1-Wasserstein distance (also known as the earth mover's distance) measures the total cost of moving shifting the mass from one probability distribution to another. For discrete probability distributions over a general metric space this can be computed exactly in $O(n^3)$ time using the Hungarian algorithm~\cite{peyre_computational_2019}, and approximated using entropic regularization in $O(n^2)$ time~\cite{cuturi_sinkhorn_2013}. However, for discrete probability distributions over a tree metric space the 1-Wasserstein distance can be computed exactly in linear time~\cite{le_tree-sliced_2019}. Given two probability distributions $\mu, \nu$ over a measurable space $\Omega$ with metric $d(\cdot,\cdot)$, let $\Pi(\mu, \nu)$ be the set of joint probability distributions $\pi$ on the space $\Omega \times \Omega$, where for any subset $\omega \subset \Omega$, $\pi(\omega \times \Omega) = \mu(\omega)$ and $\pi(\Omega \times \omega) = \nu(\omega)$. The 1-Wasserstein between $\mu$ and $\nu$ is defined as:
\begin{equation}\label{eq:wasserstein}
    W_\rho(\mu, \nu) := \inf_{\pi \in \Pi(\mu, \nu)} \int_{\Omega \times \Omega} \rho(x, y) \pi(dx, dy).
\end{equation}

Let $\| \cdot \|_{L_\rho}$ denote the Lipschitz norm w.r.t.\ $\rho$, when $\Omega$ is separable w.r.t.\ $\rho$ and $\mu, \nu$ have bounded support, then the dual of \eqref{eq:wasserstein}, known as the  Kantorovich-Rubinstein dual, can be expressed as:
\begin{equation}\label{eq:dual}
    W_\rho(\mu, \nu) = \sup_{\| f \|_{L_\rho} \le 1} \int_\Omega f(x) \mu(dx) - \int_\Omega f(y) \nu(dy). 
\end{equation}
When $d$ is a tree metric over a rooted tree $T$, for every pair of points $x, y \in \Omega$, $\rho(x,y)$ is the total weight of the (unique) path between nodes $x$ and $y$ in $T$. Denote the edge weight associated with each node $t$ as $w_t$, and $D(t, \mu)$ as the sum of mass of $\mu$ at and below node $t$, then the Wasserstein distance between two distributions on $T$ can be expressed as:
\begin{equation}
    W_{\rho_T}(\mu, \nu) = \sum_{t \in T} w_t \left | D(t, \mu) - D(t, \nu) \right |.
\end{equation}
Previous work in demonstrated unsupervised forest constructions that approximate the Wasserstein distance when $\rho$ is the Euclidean ground metric over $\Omega \equiv \mathbb{R}^d$~\cite{indyk_fast_2003, le_tree-sliced_2019, backurs_scalable_2020}. In MURAL, we construct an unsupervised random forest over a high dimensional $\Omega$ that consists of continuous, categorical and missing variables. These trees subsequently define a distance on $\Omega$, which in turn defines a Wasserstein distance between distributions on $\Omega$, and because of the specific construction of MURAL trees, admits a simple feature importance measure described in \ref{applications}. 

\section{MURAL}\label{methods}

Next we present the MURAL algorithm for building unsupervised random forests from continuous and categorical data with missing values on healthcare data. Our code is available at https://github.com/mgerasimiuk/mural.

\subsection{Problem Formulation}
Our goal is to build a distance matrix $D$ whose $(i,j)$-th entry contains the distance $d(x_i, x_j)$ between observations $x_i$ and $x_j$. Desirable properties for $D$ are that neighbors found using $D$ have similar clinical manifestations, and moreover that $D$ can be used in a nonlinear dimensionality reduction method to create nonlinear axes corresponding to largest patient variation, and clusters that group patients by overall similarities. 

\subsection{Distinguishing Randomly versus Non-Randomly Missing Variables}

A key insight in MURAL is that healthcare data consists of variables that are intentionally missing, i.e., missing not at random (MNAR), that are a source of significant information since the data is related to unobserved patient characteristics (e.g. there appears to be higher levels of missing values for reported income in individuals with  higher income levels). However, patient data also often contains data that is missing completely at random (MCAR) or missing at random (MAR). MCAR data is when missingness does not depend on the observed or missing values, and MAR data is when missingness does not depend on the missing values but may depend on the observed values. MCAR or MAR data are usually the result of absence of documentation through the extraction, transformation and loading of clinical data~\cite{devine2013, RN623}. We note that MCAR or MAR variables can be imputed on the basis of informational redundancy with other variables using conditional probability modeling, regression or other techniques, with multiple imputation leading to the most unbiased results~\cite{greenland1995,vanbuuren2007}. However, MNAR variables cannot be imputed well as we show in Section~\ref{sec:results}.

MURAL distinguishes between MCAR variables and MNAR features for continuous variables. We may or may not have prior knowledge about which continuous variables are MCAR vs.\ MNAR. If we do not, then we can distinguish between the two cases based on Little’s test~\cite{little1988test}, which examines patterns of missingness for correlations with other variables. Little's test gives each variable a significance value for rejecting the null hypothesis that says it is missing at random. If this significance value $p \le 0.05$ then we conclude that the variable is MNAR, otherwise we deem the variable to be MCAR. 

Next, for variables that are MNAR we use imputation to fill the values in as a preprocessing step using fully conditional specification (FCS) multi-variable imputation~\cite{vanbuuren2007}. After this step, all variables are either fully imputed or MNAR. 

\subsection{Mixed-modality Variable Splitting Scheme}

\begin{figure}[h]
\centering
    \includegraphics[width=\linewidth]{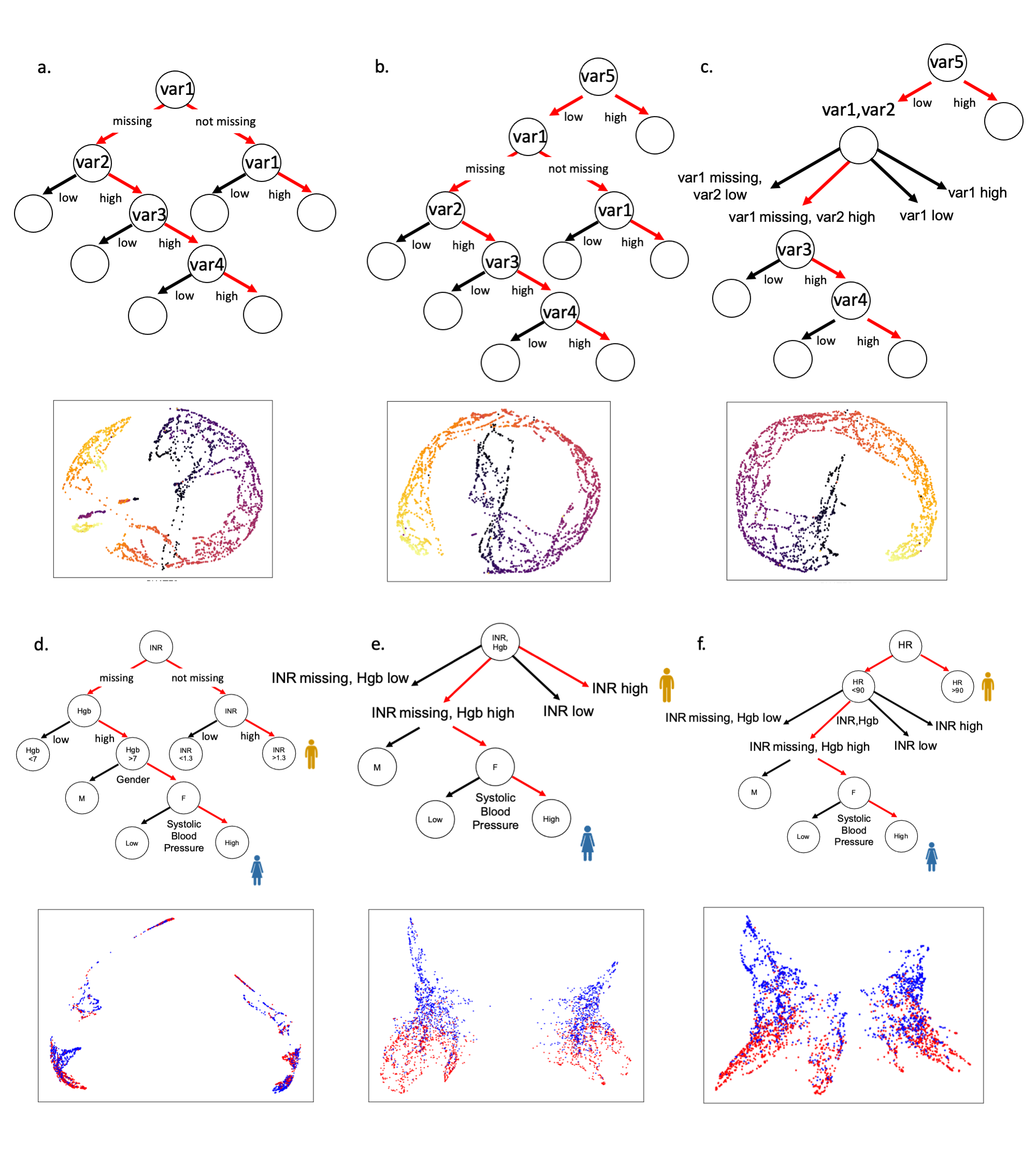}
    \caption{The decision to create the four-way split for variables missing not at random (MNAR) and to avoid splitting on MNAR variables at the root were based on empirical findings as shown above. For the 5-dimensional Swiss roll, in a) no four-way split and no condition to avoid splitting MNAR variables at the root node results in fragmentation. In b) addition of condition to avoid splitting MNAR variables at the root decreases fragmentation and c) introduction of four-way splitting results in clearer structure recovery. For the upper gastrointestinal bleeding dataset, in d) no four-way split and no-condition to avoid splitting MNAR variables at the root node results in distorted structure e) addition of four-way splitting allows for structure recovery, and f) introduction of condition to avoid splitting MNAR variables at the root leads to clearer recovery.
    \label{flatten}
}
\end{figure}

MURAL incorporates multimodal variables into an unsupervised random forest framework by using a nuanced splitting scheme.  Key aspects of the MURAL splitting scheme are as follows: 
\begin{itemize}
\item At each iteration MURAL chooses a variable $v_i$ to split on at random. 
\item If the $v_i$ is MNAR for some observations then we create a preliminary binary split between observations where it is measured and observations where it is missing. 
\item For the branch where the variable is measured, we find a single threshold based on the unsupervised information gain criterion we define in the following section to create two child nodes. 
\item For the branch where the variable is missing, we randomly select another variable $v_j$ with no missingness and create two child nodes based using the {\em residual multidimensional entropy} described in Equation \ref{eqn:infogain:split}. 
\item If $v_i$ has no missingness then it is split into two child nodes again using the same information gain criterion. 
\item The two-level splits described above are flattened into the same level to create four child nodes if the first split was missing/not-missing (see Figure \ref{flatten}). 
\end{itemize}

This scheme effectively creates four child nodes when a variable has MNAR since there are actually two variables worth of information, first variable describing the missingness, and second variable describing a further split. However the reason for choosing an additional variable on the branch with missingness is to create a branch with controlled heterogeneity (entropy) instead of the forced split that comes with a naturally binary variable. 

A similar four-way splitting scheme can avoid fragmentation caused by the presence of binary variables. For a binary variable $v_b$, we create two hidden nodes, one for the value of 0 and one for the value of 1, and choose a second, non-binary, variable $v_c$ for the hidden nodes to optimize binary splits on. The edges are weighted as in the case of missingness, resulting in a four-way split.

\subsection{Splitting with Residual Multidimensional Entropy}

When each variable without MNAR is chosen, we choose a threshold $thres(v_i)$  based on an unsupervised information gain criterion, which we term {\em residual multidimensional entropy}. Instead of choosing a split that maximizes the information gain in the class label, we choose a split that maximizes the residual entropy of the remaining variables. Since the remaining variables are not naturally discrete like class labels, we discretize the continuous variables. Thus for each node $t_i$ splitting on variable $v_i$ we compute the following summation: 

\begin{equation}
    \label{eqn:infogain:split}
    I_G(S_i) = H(S_i) -  \sum_{a \in \{L,R\}} \frac{|S_i^a|}{|S_i|} H(S_i^a)
\end{equation}

Where $S_i$ is the probability distribution of the classes (discretizations) of variable $v_i$ (not on the path from the root node to $v_i$) among $descendants (t_i)$, and $S_i^a$ is the probability distribution of the classes of $v_i$ among only the descendants of $child(t_i, a)$ i.e. the $a$th child node. We discretize each variable into a number of bins which are determined by the Sturges method~\cite{sturges1926choice} (number of bins is $\lceil \log_2 n \rceil + 1$, where n is the number of datapoints).

Here, we avoid using a high dimensional entropy or high dimensional density for ease of computation. We assume that we can approximate this entropy by a sum of marginal entropies or by the multidimensional entropy of a subset of the variables. In our experiments, we found that using three dimensional entropies calculated over randomly chosen subsets of variables resulted in the best embeddings of the Swiss roll dataset. For datasets with more dimensions (such as UGIB and eICU), three variables would cover only a small fraction of the information available, so we prefer the sum of marginal entropies. As noted before, we cannot directly use within-split affinity or density estimates as that is indeed the end result of MURAL.

\subsection{MURAL-derived Distances and Embeddings}

For every tree $T_k$ in a trained forest, we define the tree distance between two nodes to be

\begin{equation}
    \label{eqn:treedist}
    d_k(t_i, t_j) = \sum_{p \in \mathcal{P}} w_{t_p}
\end{equation}
where $\mathcal{P}$ is the index sequence of the edges in $T_k$ on the unique undirected path $(t_i, \ldots, t_j)$ between $t_i$ and $t_j$ with no node repetition, and $w_{t_p}$ is the weight of edge $p$. We opt for all edges having unit weights. 

Then we define distances between $x_i, x_j$ by noting that nodes corresponding to leaves of the tree $t_{l,1}, t_{l,2}, \ldots $ each contain sets of datapoints, i.e., $t_{l,1}= \{x_{l,1}, x_{l,2} \ldots x_{l,m}\}$. Thus if $x_i \in t_{l,i}$ and $x_j \in t_{l,j}$ then $d_k(x_i, x_j) = d_k(t_{l,i}, t_{l,j})$

The constructed tree metrics can be averaged over the MURAL-forest, which results in a composite MURAL-distances.  

\begin{equation}
    \label{eqn:muraldist}
    D_M = (1/k)\sum_k D_k.
\end{equation}
Here $D_k$ is the matrix of pairwise distances of the $k$th tree, and $D_M$ the the averaged distance matrix over all trees. 

This leads directly to a MURAL-based distance matrix, which can be converted into an affinity matrix using any kernel function. For embedding, the distance matrix can be passed as an input to a nonlinear dimensionality reduction algorithm such as PHATE \cite{Moon19}, which we choose for its manifold-affinity preservation capabilities. We call the resultant PHATE embedding using $D_M$ the {\em MURAL-embedding}. 

\section{Results}
\label{sec:results}

\subsection{Empirical Validation}
\label{datasets}

In this section we induced missingness, i.e., MNAR, in datasets (or data subsets) where there was no missingness to validate the ability of MURAL to recover meaningful distances. We used 3 datasets for this experiment:

\begin{enumerate}
    \item A synthetic Swiss roll constructed of 3,000 points embedded in a 5 dimensional space. 
    \item  A dataset of patients requiring intensive care unit stay (ICU) from the publicly available Phillips eICU Collaborative Research Database (eICU-CRD) of critical care units across the United States from 2014 to 2015~\cite{eicu_physionet,eicu_2}. Patients with available data within the first 24 hours of ICU stay (148,532 unique patient encounters) had 10 variables selected: five laboratory values (bilirubin, blood urea nitrogen, creatinine, hematocrit, and albumin), age, 3 ordinal variables from nursing assessment of eye, motor, and verbal responses using the Glasgow Coma Scale, and binary variable of invasive mechanical ventilation within the first 24 hours. 
    \item An international registry of consecutive, unselected patients presenting with symptoms of upper gastrointestinal bleeding between March 2014 and March 2015 from Yale–New Haven Hospital (United States), Glasgow Royal Infirmary (Scotland), Royal Cornwall Hospital Truro (England), Odense University Hospital (Denmark), Singapore General Hospital (Singapore), and Dunedin Hospital (New Zealand)~\cite{Stanleyi6432}. 7 variables were selected: 4 laboratory variables (hemoglobin, urea, albumin, INR), 1 vital sign (systolic blood pressure), 1 binary demographic variable (gender), and 1 ordinal variable (degree of liver disease).
\end{enumerate}
The eICU and UGIB datasets already have missingness in a significant portion of the entries. Thus, to create an artificial ground truth,  we used only entries with all variables present. For the UGIB dataset, 2,761 patients with complete data were selected. For computational efficiency we subsampled the eICU dataset to 10,000 patient encounters with complete data across 10 variables. Using this subset of complete data as ground truth, we artificially induced missingness in order to test the ability of MURAL to recover meaningful distances.   

We induced missingness in the Swiss roll dataset in a similar way to what is observed in real clinical data, with one variable that has a pattern of missing values deemed to be missing not at random by pairwise Little's test~\cite{little1988test}, and two variables with random values dropped out. In the eICU dataset we induced missingness by dropping the values of bilirubin $>3$, the threshold chosen since these are clearly physiologically abnormal values and thus would be missing not at random. In the UGIB dataset we induced missingness by dropping the values of INR $>3$, also chosen since they are physiologically abnormal values and would be missing not at random. All datasets were standardized after missingness was induced but before they were used for constructing MURAL-forests.

To quantitatively compare the preservation of the underlying manifold structure of these datasets in the presence of missing values, we used the accuracy of a kNN graph derived from the embedding compared to the ground truth kNN graph from each of these graphs.  We compared the performance to mean imputation, and to another standard tree-based imputation method, multiple imputation with chained equations (MICE) using classification and regression decision trees. We find that near neighbors are recovered in the MURAL-embedding with greater accuracy than baselines that first use imputation in this case where ground truth is known in \ref{table1}. In the artificial Swiss roll dataset, the MURAL-forest with 100 trees outperformed mean imputation at 5, 10, and 100 neighbors by 32\%, 30\%, and 22\%, and outperformed MICE with CART at 5 and 10 neighbors by 6\% and 3\%. This reflects the specific nature of the Swiss roll dataset, since smaller neighborhoods were more likely to be perturbed due to the intrinsic coiled data structure manifold. For the eICU dataset and the UGIB dataset, the MURAL-forest with 100 trees outperformed mean imputation at 5, 10, and 100 neighbors by 7\%, 10\%, and 15\%; MURAL-forest outperformed MICE with CART at 5, 10, and 100 neighbors by 15\%, 15\%, and 14\%. 

\begin{table}[tb]
\label{table1}
\caption{
$\mu \pm \sigma$ for P@5, 10, 100 metrics on three datasets over 5 runs. MURAL-embedding preserves neighborhoods for missing values better than mean imputation and MICE with CART.}
\begin{center}
\begin{tabular}{lrrr} 
    \toprule
    {} &   P@5 & P@10 & P@100\\
    \midrule
    Swiss Roll &  {} &  {}  &  {}  \\ 
    MURAL &  \textbf{0.729 $\pm$ 0.01}  &  \textbf{0.752 $\pm$ 0.01}  &  0.743 $\pm$ 0.01  \\
    Mean Imputation  &  0.403  $\pm$ 0.00 &  0.429$\pm$ 0.00  &  0.522 $\pm$ 0.00 \\
    MICE with CART  &  0.664 $\pm$ 0.01  &  0.729 $\pm$ 0.01  &  \textbf{0.850 $\pm$ 0.01} \\
    \midrule
    eICU dataset &  {}  &  {}  &  {} \\
    MURAL &  \textbf{0.227 $\pm$ 0.01}  &  \textbf{0.259 $\pm$ 0.02}  &  \textbf{0.387 $\pm$ 0.06} \\
    Mean Imputation  &  0.158 $\pm$ 0.00  &  0.162 $\pm$ 0.00 &  0.231 $\pm$ 0.00\\
    MICE with CART  &  0.119 $\pm$ 0.01  & 0.135 $\pm$ 0.01  &  0.215 $\pm$ 0.00 \\
   \midrule
    UGIB dataset &  {}  &  {}  &  {}\\
    MURAL &  \textbf{0.239 $\pm$ 0.02}  &  \textbf{0.230 $\pm$ 0.03}  &  \textbf{0.307 $\pm$ 0.03}\\
    Mean Imputation  &  0.080 $\pm$ 0.00 &  0.080 $\pm$ 0.00 &  0.150 $\pm$ 0.00\\
    MICE with CART &  0.085 $\pm$ 0.01  &  0.084 $\pm$ 0.01 &  0.165 $\pm$ 0.01\\
    \bottomrule
\end{tabular}
\end{center}
\end{table}

\begin{table*}[tb]
\caption{Ablation Study for MURAL-embedding on the Swiss Roll Embedded in 5 Dimensions. $(\mu \pm \sigma)$ over 5 runs. \textbf{Bold} represents best in each parameter.}
\label{ablation1}
\begin{center}
\begin{tabular}{lrrrrrrr}
 \toprule
 Model & DeMAP & P@5 & P@10 & P@100 & P@500 & Distortion & Time (s)\\ 
 \midrule
 Mean Imputation & 0.495 & 0.637 & 0.635 & 0.581 & 0.675 & 5530 & ---\\
 \midrule
 \textbf{Entropy Dimensions} & {}&{}& {}&	{}&	{} & {}& {}\\
 1D & 0.504 $\pm$ 0.08 & 0.720 $\pm$ 0.01 & 0.742 $\pm$ 0.01 & 0.739 $\pm$ 0.01 & \textbf{0.783 $\pm$ 0.01} & 1880 $\pm$ 270 & 376 $\pm$ 240\\
 2D & 0.618 $\pm$ 0.04 & 0.728 $\pm$ 0.01 & 0.748 $\pm$ 0.01 & 0.746 $\pm$ 0.01 & 0.782 $\pm$ 0.00 & 1800 $\pm$ 180 & \textbf{150 $\pm$ 5.4}\\
 \textbf{3D}  & 0.577 $\pm$ 0.08& 0.733 $\pm$ 0.01 & 0.755 $\pm$ 0.01 &	\textbf{0.749 $\pm$ 0.01} & 0.779 $\pm$ 0.01 & 1860 $\pm$ 250 & 240 $\pm$ 22\\
 5D & \textbf{0.628 $\pm$ 0.03} & 0.704 $\pm$ 0.01 & 0.725 $\pm$ 0.00 & 0.727 $\pm$ 0.02 & 0.763 $\pm$ 0.01 & $1.23 \pm 2.0 \times 10^7$ & 1150 $\pm$ 260\\
 RME & 0.556 $\pm$ 0.06 & \textbf{0.738 $\pm$ 0.01} & \textbf{0.757 $\pm$ 0.01} & 0.738 $\pm$ 0.01 & 0.773 $\pm$ 0.01 & \textbf{1540 $\pm$ 120} & 448 $\pm$ 41\\
 \midrule
 \textbf{\# Variables Split} & {}&{}& {}&	{}&	{} & {} &{}\\
 \textbf{1} & \textbf{0.582 $\pm$ 0.08} & 0.733 $\pm$ 0.00 &	\textbf{0.755 $\pm$ 0.01} & \textbf{0.749 $\pm$ 0.01} & \textbf{0.779 $\pm$ 0.01} & \textbf{1860 $\pm$ 250} & \textbf{286 $\pm$ 23}\\
 2  & 0.542 $\pm$ 0.07 & 0.752 $\pm$ 0.01 &	0.781 $\pm$ 0.01 & 0.725 $\pm$ 0.01 & 0.751 $\pm$ 0.01 & 2150 $\pm$ 81 & 517 $\pm$ 5.7\\
 3  & 0.550 $\pm$ 0.05 & 0.757 $\pm$ 0.01 &	0.782 $\pm$ 0.01 & 0.721 $\pm$ 0.01 & 0.741 $\pm$ 0.01 & 2020 $\pm$ 200  & 989 $\pm$ 110\\
 4  & 0.530 $\pm$ 0.06& \textbf{0.762 $\pm$ 0.00} &	0.781 $\pm$ 0.00 & 0.711 $\pm$ 0.01 & 0.738 $\pm$ 0.01 & 2140 $\pm$ 180 & 1240 $\pm$ 140\\
 \midrule
 \textbf{Restrict MNAR Levels} & {}&{}& {}&	{}&	{} & {} &{}\\
 0  & 0.552 $\pm$ 0.05 & 0.712 $\pm$ 0.01 & 0.725 $\pm$ 0.01& 0.573 $\pm$ 0.02& 0.538 $\pm$ 0.03 & 2180 $\pm$ 360 & 274 $\pm$ 33\\
 1 & 0.604 $\pm$ 0.04 & 0.722 $\pm$ 0.01 & 	0.728 $\pm$ 0.00 & 0.703 $\pm$ 0.00 & 0.744 $\pm$ 0.01 & 2060 $\pm$ 230 & 297 $\pm$ 22\\
 2 & \textbf{0.636 $\pm$ 0.03} & 0.716 $\pm$ 0.01 & 	0.730 $\pm$ 0.01 & 0.737 $\pm$ 0.01 & 0.772 $\pm$ 0.00 & 1920 $\pm$ 170 & 304 $\pm$ 16\\
 \textbf{3} & 0.570 $\pm$ 0.08& \textbf{0.733 $\pm$ 0.00}& \textbf{0.755 $\pm$ 0.01}& \textbf{0.749 $\pm$ 0.01 }& \textbf{0.779 $\pm$ 0.01} & \textbf{1860 $\pm$ 250} & \textbf{264 $\pm$ 34}\\
  \midrule
 \textbf{Tree Depth} & {}&{}& {}&	{}&	{} & {} & {}\\
  2 & 0.189 $\pm$ 0.01 & 0.477 $\pm$ 0.02 & 0.527 $\pm$ 0.02 & 0.609 $\pm$ 0.01 & 0.721 $\pm$ 0.02 & $8.61 \pm 3.5 \times 10^8$ & \textbf{55.7 $\pm$ 1.6}\\
  4 & 0.259 $\pm$ 0.02 & 0.655 $\pm$ 0.01 & 0.688 $\pm$ 0.01 & 0.685 $\pm$ 0.01 & 0.769 $\pm$ 0.01 & $7.17 \pm 1.3 \times 10^7$ & 110 $\pm$ 8.7\\
  6 & 0.614 $\pm$ 0.05 & 0.707 $\pm$ 0.01 & 0.744 $\pm$ 0.01 & 0.722 $\pm$ 0.01 & 0.780 $\pm$ 0.01 & $5.27 \pm 4.9 \times 10^6$ & 164 $\pm$ 10\\
  8 & 0.623 $\pm$ 0.05 & 0.724 $\pm$ 0.01 & 0.746 $\pm$ 0.01 & \textbf{0.747 $\pm$ 0.01} & \textbf{0.784 $\pm$ 0.01} & 1950 $\pm$ 130 & 194 $\pm$ 16\\
  \textbf{10}& 0.624 $\pm$ 0.03 & \textbf{0.733 $\pm$ 0.01} & \textbf{0.753 $\pm$ 0.01} &	0.743 $\pm$ 0.01 & 0.776 $\pm$ 0.01 & \textbf{1790 $\pm$ 67} & 174 $\pm$ 26\\
  12& 0.623 $\pm$ 0.01 & 0.729 $\pm$ 0.01 & 0.750 $\pm$ 0.01 &	0.744 $\pm$ 0.01 & 0.773 $\pm$ 0.01 & 1940 $\pm$ 110 & 195 $\pm$ 1.3\\
  14& \textbf{0.646 $\pm$ 0.03} & 0.729 $\pm$ 0.01 & 0.751 $\pm$ 0.01 &	0.745 $\pm$ 0.01 & 0.774 $\pm$ 0.01 & 1833 $\pm$ 180 & 200 $\pm$ 5.8\\
   \midrule
 \textbf{Forest Size} & {}&{}& {}&	{}&	{} & {} & {}\\
  10 & 0.368 $\pm$ 0.13 & 0.597 $\pm$ 0.01 & 0.647 $\pm$ 0.01 & 0.702 $\pm$ 0.2 & 0.723 $\pm$ 0.02 & $1.39 \pm 3.9 \times 10^8$ & \textbf{41 $\pm$ 43}\\
  50 & 0.596 $\pm$ 0.07 & 0.708 $\pm$ 0.01 & 0.734 $\pm$ 0.01 & 0.737 $\pm$ 0.01 & 0.765 $\pm$ 0.01 & $5.94 \pm 8.0 \times 10^6$ & 108 $\pm$ 40\\
  \textbf{100} & 0.624 $\pm$ 0.03 & 0.733 $\pm$ 0.01 & 0.753 $\pm$ 0.01 &	0.743 $\pm$ 0.01 & 0.776 $\pm$ 0.01 & \textbf{1790 $\pm$ 67} & 174 $\pm$ 26\\
  200& \textbf{0.632 $\pm$ 0.03} & 0.746 $\pm$ 0.01 & 0.763 $\pm$ 0.01 &	0.755$\pm$ 0.01 & 0.785 $\pm$ 0.01 & 1852 $\pm$ 120 & 332 $\pm$ 31\\
  500 & 0.622 $\pm$ 0.03 & \textbf{0.749 $\pm$ 0.01} & \textbf{0.769 $\pm$ 0.01} &	\textbf{0.762 $\pm$ 0.00} & \textbf{0.791 $\pm$ 0.00} & 1850 $\pm$ 61 & 801 $\pm$ 83\\
 \bottomrule
\end{tabular}
\end{center}
\end{table*}

\begin{figure}[h]

\centering
    \includegraphics[width=9cm]{"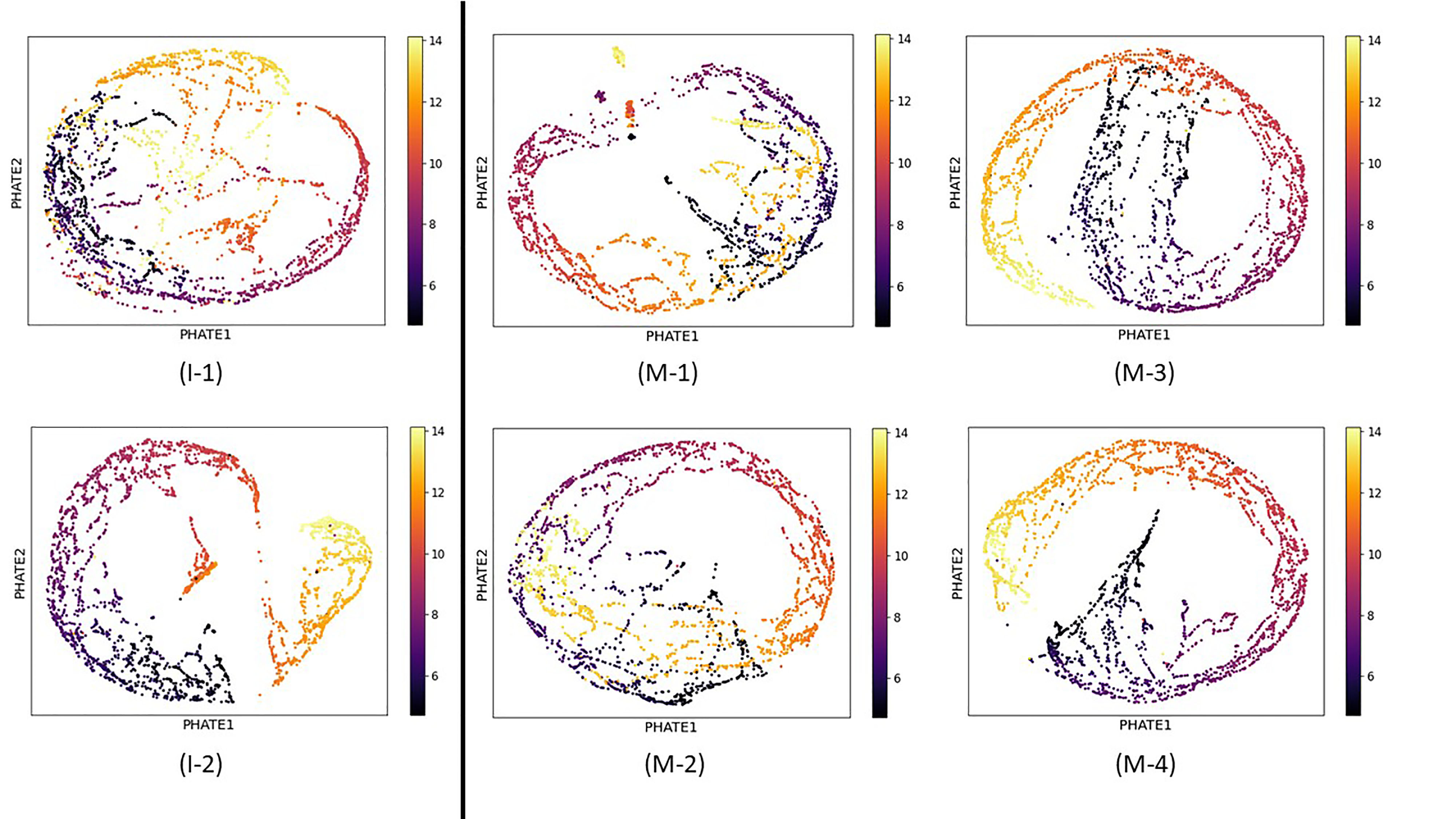"}
    \caption{PHATE plots of the Swiss roll embedded in 5 dimensions. (I-1) Mean imputation. (I-2) MICE. (M-1) MURAL-embedding without any restrictions on choosing splits in variables with missing values. (M-2) MURAL-embedding with trees of depth 4. (M-3) MURAL-embedding with each node choosing the best split from among four variables. (M-4) MURAL-embedding with 100 trees of depth 10, each node choosing the best split from only one variable, not splitting on variables with missing values in the first three levels.}
    \label{swissroll}
\end{figure}

\begin{figure*}[!t]
\centering
    \includegraphics[width=\linewidth]{"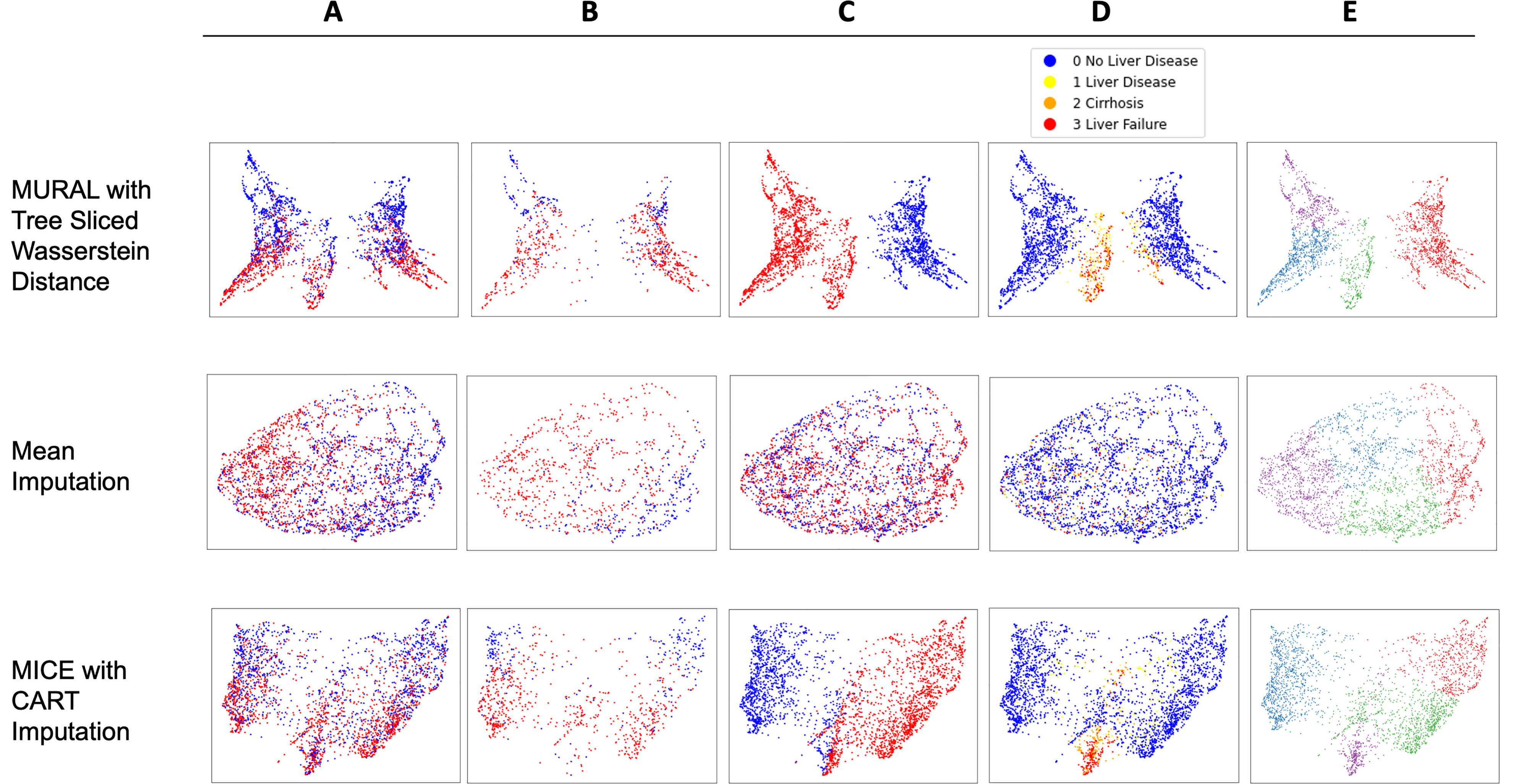"}
    \caption{MURAL-embedding preserves data structure and separation better than  mean imputation and multiple imputation with chained equations using classification and regression trees (MICE with CART) after visualization with PHATE. MURAL-embedding separates groups of patients with clinically relevant subgroups: A) high risk (red) and low risk (blue) groups as defined by need for hospital-based intervention B) different age groups $<$30 years old (blue) versus $>$80 years old (red) C) gender, male (red) versus female (blue) D) liver disease (yellow to red) versus no liver disease (blue). Spectral clustering (E) of the graph subsets the known groups including the group of men with liver disease (green), female patients (red), young males at low risk for hospital-based intervention (purple), and older males at high risk for hospital-based intervention (blue).}
    \label{clustersUGIB}
\end{figure*}

\begin{figure*}[!t]
\centering
    \includegraphics[width=\linewidth]{"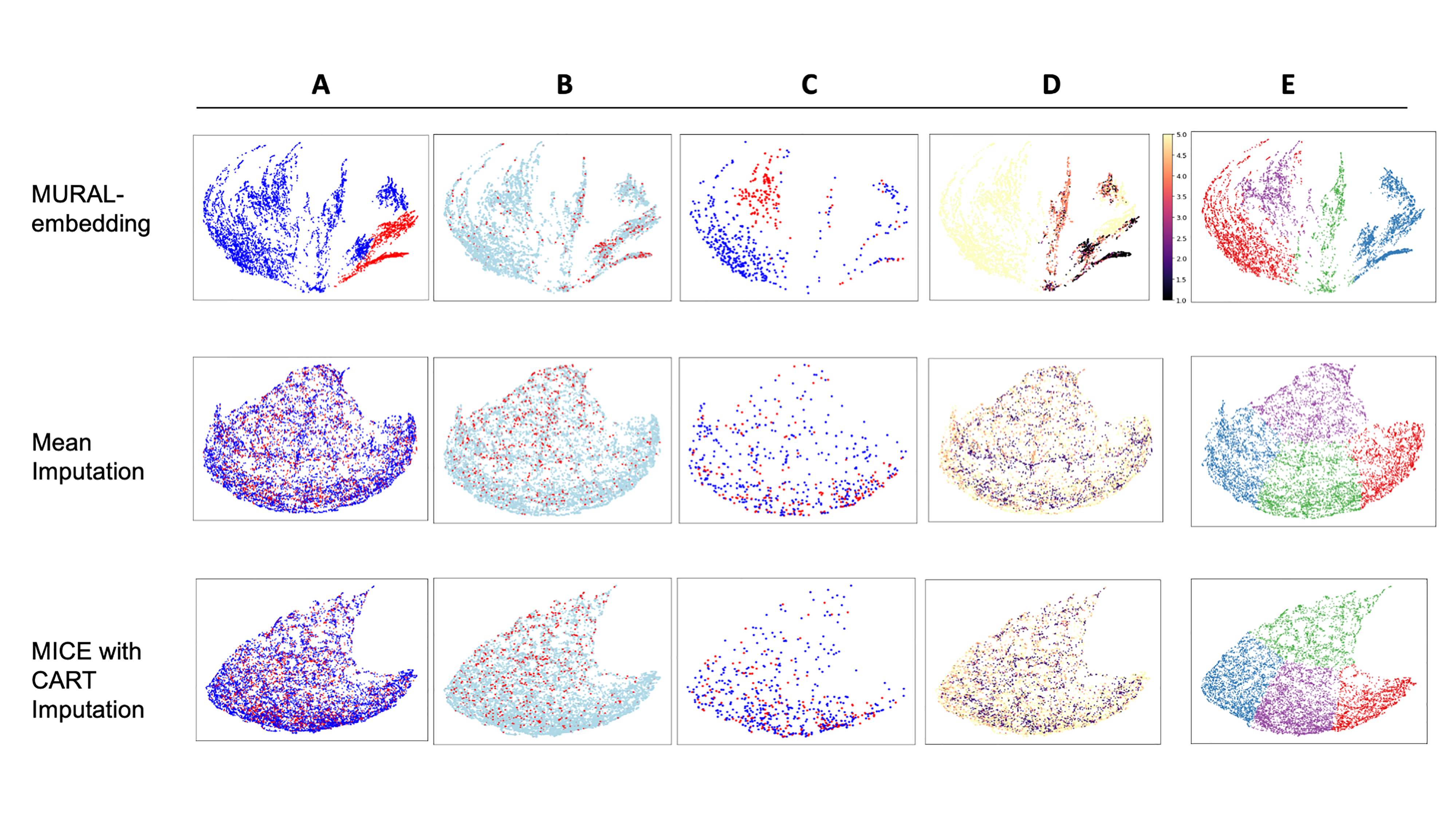"}
    \caption{MURAL-embedding better separates out patients in the intensive care unit at risk for in-hospital mortality compared to mean imputation and MICE with CART on PHATE visualizations. MURAL-embedding separates patients who A) received invasive mechanical ventilation (red) versus not (blue) B) at high risk for in hospital mortality (red) versus not (light blue) C) presenting with admission diagnosis of acute coronary syndrome (red) versus other diagnosis (blue) D) nursing assessment (1 to 5, higher is healthier). Spectral clustering (E) of the graph subsets the groups of patients who required mechanical ventilation (blue), patients with slightly impaired verbal responses on nursing assessment (green), and patients with admission diagnosis of acute coronary syndrome with bilirubin measurement, reflecting concern about liver disease (purple) versus without bilirubin measurement (red), impairment of verbal responses (blue).}
    \label{clusterseICU}
\end{figure*}

\begin{figure*}[!t]
\centering
    \includegraphics[width=\linewidth]{"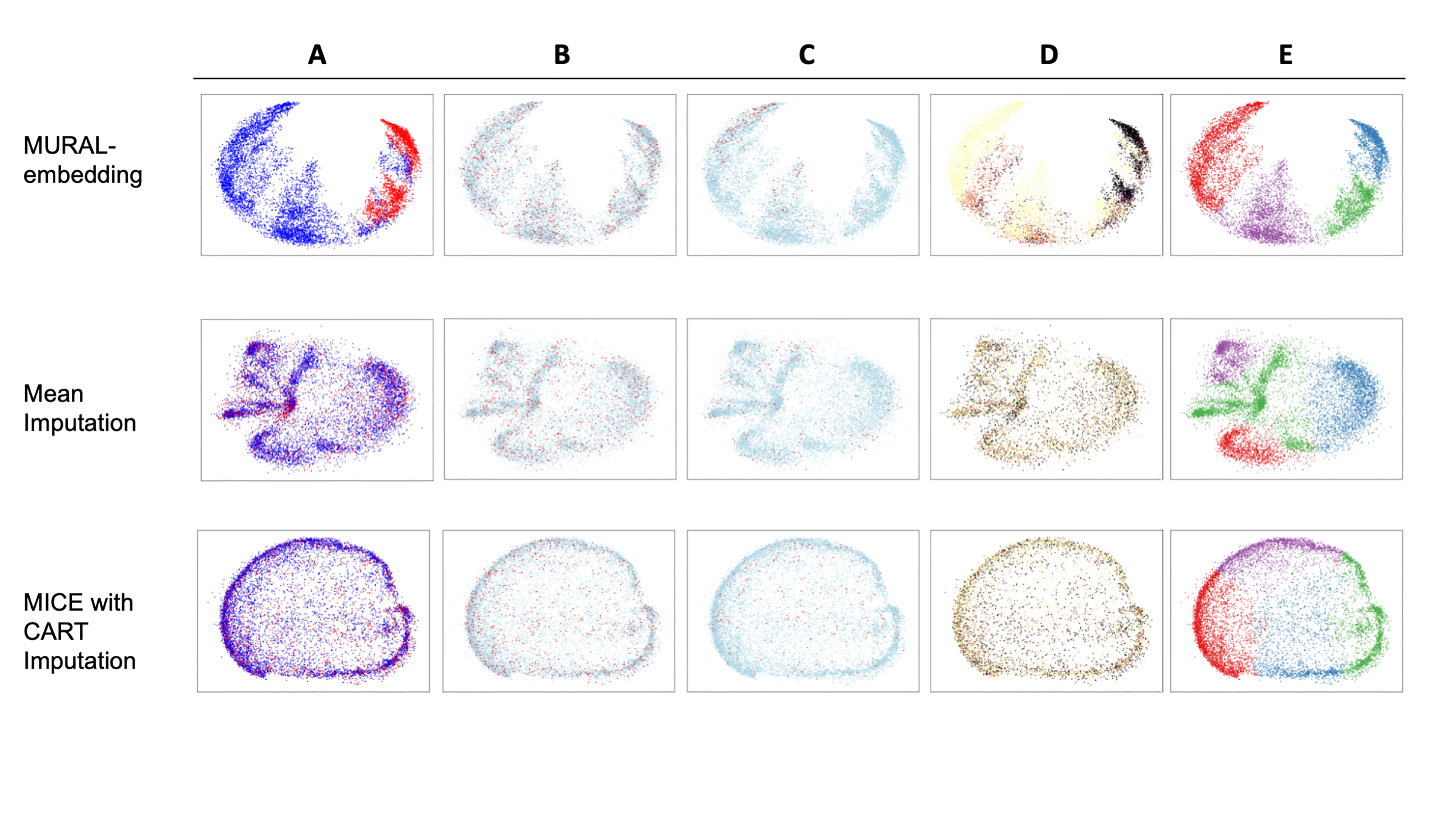"}
    \vspace{-18mm}
    \caption{MURAL-embedding better separates out patients in the intensive care unit at risk for in-hospital mortality compared to mean imputation and MICE with CART on PHATE visualizations even with 108 variables. MURAL-embedding separates patients who A) received invasive mechanical ventilation (red) versus not (blue) B) at high risk for in hospital mortality (red) versus not (light blue) C) presenting with admission diagnosis of acute coronary syndrome (red) versus other diagnosis (blue) D) nursing assessment (1 to 5, higher is healthier). Spectral clustering (E) of the graph subsets the group of patients who required mechanical ventilation (blue/green)}
    \label{clusterseICU_2}
\end{figure*}

\begin{table}[tb]
\caption{MURAL-embedding on PHATE with spectral clusters k=4 shows superior silhouette scores compared to other imputation methods}
\label{silhouette}
\begin{center}
\begin{tabular}{llr} 
    \toprule
    Experiment & Approach & Silhouette Score \\
    \midrule
    UGIB dataset & MURAL & \textbf{0.46}  \\
    {} & Mean Imputation & 0.043 \\
    {} & MICE with CART  & 0.44 \\
     \hline
    eICU dataset & MURAL & \textbf{0.390}  \\
    {} & Mean Imputation & 0.386 \\
    {} & MICE with CART  & 0.378 \\
    \bottomrule

\end{tabular}
\end{center}
\end{table}

\begin{figure*}[!t]
\centering
    \includegraphics[width=\linewidth]{"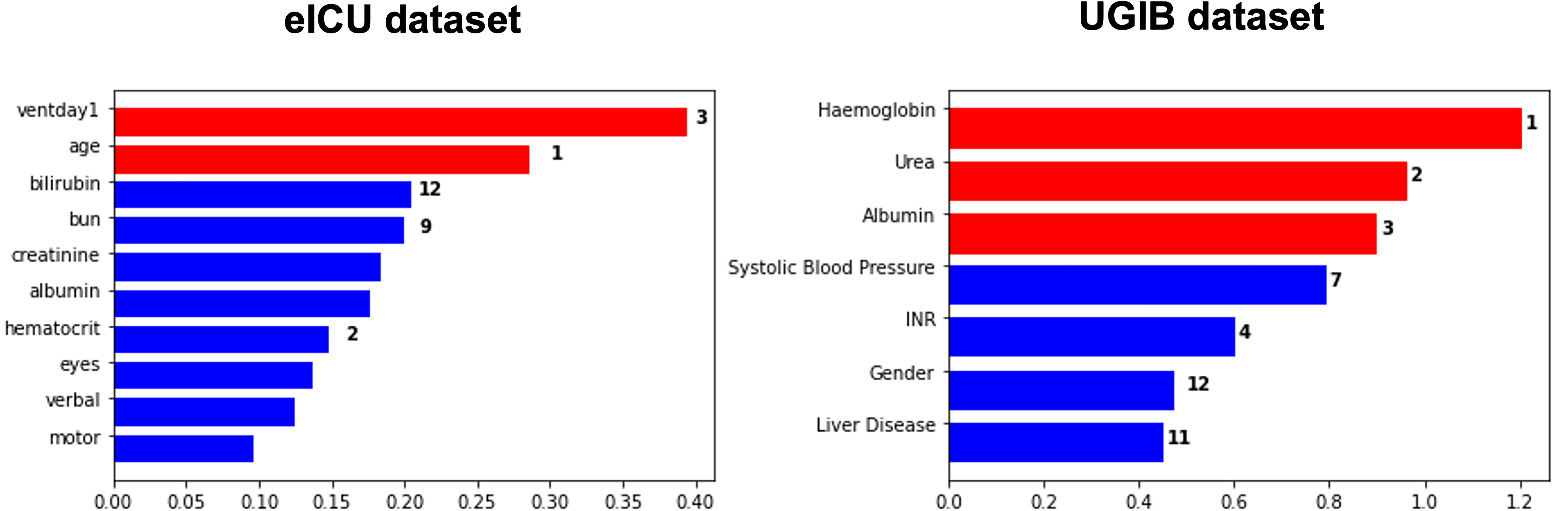"}
    \caption{TSWD on the MURAL-forest can be used to generate feature importance graphs that are consistent with feature importances in supervised approaches on the same data. For the eICU dataset, the first two factors were within the top 3 predictive factors in supervised models predicting mortality on the same dataset. For the UGIB dataset, the first 3 factors were identified as the top 3 predictive factors in high-performing supervised models on the same dataset.}
    \label{feature_imp}
\end{figure*}

\begin{table*}[tb]
\caption{Tree Sliced Wasserstein Distances (TSWD) on the MURAL-forest compared to mean imputation and multiple imputation with chained equations using classification and regression trees (MICE with CART). TSWD on the MURAL-forest appear to separate populations by clinically meaningful risk (either in-hospital mortality for patients in the ICU or need for hospital based intervention for patients with UGIB) more definitively than the other approaches, with an increased ratio of Earth Mover's Distance (EMD) of the two defined populations to EMD between random splits. Risk based on need for hospital-based intervention (UGIP) and in-hospital mortality (eICU).}
\begin{center}
\begin{tabular}{llrrrr} 
    \toprule
    Experiment & Approach & Risk & $<$30 y/o v $>$80 y/o & $<$30 y/o v 30-40 y/o\\
    \midrule
    UGIB dataset & MURAL-forest with TSWD &   \textbf{12.0 $\pm$ 0.33}  &  12.6 $\pm$ 0.56  &  5.2 $\pm$ 0.22  \\
    {} & EMD on Mean Imputation &  2.39  &  2.26  &  1.27\\
    {} & EMD on MICE with CART  &  1.13 $\pm$ 0.001  &  1.12 $\pm$ 0.001  &  0.94 $\pm$ 0.0001 \\
     \midrule
    eICU dataset & MURAL-forest with TSWD & \textbf{10.6 $\pm$ 1.76}  &  28.5 $\pm$ 5.8  &  1.96 $\pm$ 0.44  \\
    {} & EMD on Mean Imputation &  1.78  &  3.78  &  1.82\\
    {} & EMD on MICE with CART  &  1.73 $\pm$ 0.01  &  3.60 $\pm$ 0.01  &  1.68 $\pm$ 0.08 \\
    \bottomrule

\end{tabular}
\end{center}
\end{table*}

\subsection{Ablation Study}

Next, we carried out an ablation study using this Swiss roll dataset to investigate which parameter choices lead to the best embeddings. Results are shown in Table~\ref{ablation1}. For each parameter choice, we trained MURAL-forests with 5 different random initializations. In the run testing 100 trees of depth 10, we rejected one outlier forest that generated pathological distances. Generally we find that low-dimensional (3-d) entropy works for our proposed residual multidimensional entropy from Equation~\ref{eqn:infogain:split}. Surprisingly we found that splitting on a {\em single} residual variables works best for the discrete and MNAR case. An additional insight was that restricting MNAR variables to low levels in the tree worked best as they would have minimal effect on other distances in this scheme. Full results are seen in Table~\ref{ablation1}. 

\subsection{Applications}
\label{applications}

\subsubsection{Embeddings}

Our initial goal was to derive distances that provide a faithful representation of a dataset that contains different types of variable and missing values. In order to see if these distances facilitate embeddings that reveal structure and meaningful groupings in data, we fed the distance matrix into the PHATE \cite{Moon19} nonlinear dimensionality reduction and visualization method. We chose to use PHATE due to its improved ability to preserve data manifold-affinities as quantified by the DeMAP metric in \cite{Moon19}. We note that PHATE contains similar information as a diffusion map \cite{coifman:diffusionMaps2006}, with information collected in low dimensions for visualization.  The resultant embeddings, which we term as MURAL-embeddings are shown in Figure~\ref{swissroll}, Figure~\ref{clustersUGIB} and Figure~\ref{clusterseICU}. 

By visual inspection we see that the MURAL-embeddings are much more structured than PHATE embeddings of the raw data with imputed values. Furthermore, visualization of several clinical variables on the embeddings show that the separations correspond to clinical groupings that are used in generating the embedding and also meaningful clinical groups not used to generate the MURAL-embeddings. In Figure~\ref{clustersUGIB} risk for hospital-based intervention (A) and age (B) were not used to generate the MURAL-embeddings, yet the MURAL-embeddings show separation into high risk and low risk groupings (A) and separation by age, in this case $<30$ years old versus $>80$ years old (B). For factors used to generate the MURAL embeddings, there are separations into two major structures horizontally based on gender (C) and minor structures based on liver disease status (D). By contrast, mean imputation mixes genders despite having the information, and MICE imputation overlaps the two disease cohorts. In Figure \ref{clusterseICU} risk for in-hospital mortality (B) and admission diagnosis of acute coronary syndrome (C) were not used to generate the MURAL-embeddings, but the MURAL-embeddings show separation into low and high risk groups for in-hospital mortality (B) and subgroup of patients with admission diagnosis of acute coronary syndrome (C) (red) and those with other diagnoses (blue). Need for mechanical ventilation (A) and nursing assessment of verbal response (D) were used to generate the MURAL-embedding; clear separations are found between patients with mechanical ventilation (A, red) and those who did not (A, blue), as well as patients with normal verbal response (D, light yellow), mildly impaired verbal response (D, orange), those with no verbal response (D, purple). These separations are not seen in the embedding methods on imputed data despite also containing these variables. These visual results suggest that these embeddings are useful and amenable to further quantitative analysis. The structures in the MURAL-embedding are largely retained even after including 108 variables derived from the first 24 hours of patient stay in the ICU for the same patient cohort in the eICU dataset (Figure \ref{clusterseICU_2}).

\subsubsection{Spectral Clustering}

Spectral clustering \cite{ng2001spectral} is a key unsupervised clustering method that follows the data manifold by operating on a data affinity matrix. We used k-means with $k=4$ on the diffusion operator created by PHATE, which is equivalent to spectral clustering and compared it to similar clustering using imputed data. To evaluate cluster quality we use the silhouette score, and found that the clustered MURAL-embedding has highest silhouette score in both datasets. 

In addition, the resultant subgroups can be interpreted clinically. For example, the red cluster in Figure \ref{clusterseICU} corresponds to admission diagnosis of acute coronary syndrome who either have a bilirubin measured or not measured within 24 hours of ICU admission, which may suggest a concern for concurrent hepatobiliary dysfunction or injury. In Figure~\ref{clusterseICU} the green cluster correspond to male patients with severe liver disease, which correspond to a specific type of gastrointestinal bleeding, portal hypertensive bleeding. 

\subsubsection{Wasserstein Distance between cohorts}

Seeing that groups of patients who differ in a clinically significant way form distinct clusters on the embeddings, we decided to quantitatively check whether MURAL-forests themselves separate dissimilar cohorts more than similar ones. This is directly relevant to the task of characterizing clusters within the data representations. For example, if clusters are two groups of patients with the same diagnosis, if there is a meaningful difference between their measured laboratory or clinical characteristics that could reflect risk for a poor clinical outcome. To that end, we calculated tree-sliced Wasserstein distances between the low risk and high risk patient cohorts. For the eICU dataset, we defined this as patients who died in the hospital, and in the UGIB dataset we defined this as a need for hospital-based intervention (red blood cell transfusion, hemostatic intervention, or 30-day mortality). 
We then extract feature importances by aggregating the variables associated with nodes where the Wasserstein distances were most disparate. The feature importances in Figure~\ref{feature_imp} are consistent with the top two variables predictive of in-hospital mortality from supervised machine learning algorithms (regression models) trained and validated on the eICU dataset~\cite{cosgriff}, and the top three variables predictive of need for hospital based intervention from a high performing supervised machine learning algorithm (gradient boosted decision trees) trained and validated on the UGIB dataset~\cite{SHUNG2020160}. As a sanity check, we also compared tree-sliced Wasserstein distances between different age groups: first, a very different age group ($<30$ years old versus $>80$ years old) and a similar age group ($<30$ years old versus 30-40 years old). The similar age group had much lower TSWD compared to the group with different ages, and the TSWD from the MURAL-forest for high risk versus low risk was more distinct compared to the other imputation methods. 
More generally, we believe these types of Wasserstein distances can be used to measure distances or similarities between treatments, diagnostic variants and other differences between sub-cohorts. 

\section{Conclusion}
We present MURAL, a random-forest based framework for deriving distances between patients using mixed-model electronic health record data. We showed that the resultant MURAL-embeddings recapitulate the structure and heterogeneity of patient populations better than alternatives---thus paving the way for unsupervised learning to be used on clinical data. We note that most of the machine learning methods that are currently used for modeling clinical data require supervised training and large sets of annotated and labeled samples. However, by making clinical data amenable to unsupervised approaches, we can diminish this burden and even discover novel, clinical groupings  of patients that could be meaningful for diagnosis, prognosis, or treatment. 

\clearpage
\bibliographystyle{ieeetr}
\bibliography{bibliography.bib}

\end{document}